[Preprint version]

# A Spatio-Temporal Graph Neural Networks Approach for Predicting Silent Data Corruption inducing Circuit-Level Faults

Shaoqi Wei*

Ehime University, Japan, k819002y@mails.cc.ehime-u.ac.jp

Senling Wang

Ehime University, Japan, wang@mails.cc.ehime-u.ac.jp

Hiroshi Kai

Ehime University, Japan, kai@cs.ehime-u.ac.jp

Yoshinobu Higami

Ehime University, Japan, higami@cs.ehime-u.ac.jp

Ruijun MA

Anhui University of Sci.&Tech., China, ruijun_ma@foxmail.com

Tianming NI

Anhui Polytechnic University, China, timmyni126@126.com

Xiaoqing WEN

Kyushu Institute of Technology, Japan, wen@csn.kyutech.ac.jp

Hiroshi Takahashi

Ehime University, Japan, takahashi@cs.ehime-u.ac.jp

Silent Data Errors (SDEs), commonly caused by time-zero defects and aging degradation, undermine the reliability of safety-critical computing systems. Functional testing helps detect SDE-inducing faults, but it suffers from high simulation costs. We propose a unified Spatio-Temporal Graph Convolutional Network (ST-GCN) framework that enables efficient and accurate long-cycle Fault Impact Probability (FIP) prediction for large-scale sequential circuits, thereby providing critical support for the quantitative assessment and risk evaluation of functionally possible faults. By representing gate-level netlists as spatio-temporal graphs, our approach captures both the circuit topology and the temporal dynamics of signal propagation. The ST-GCN leverages dedicated spatial and temporal encoders to extract comprehensive features and efficiently predict multi-cycle FIPs. Experimental results on ISCAS'89 benchmarks show that our method reduces simulation time by over an order of magnitude while achieving high prediction accuracy (mean absolute error as low as 0.024 for 5-cycle predictions). Moreover, the framework accommodates both testability-metric-based and fault-simulation-based feature modeling, enabling flexible trade-offs between computational efficiency and accuracy. We further validate the practical effectiveness of ST-GCN through a test point selection case study, where optimizing observation points based on predicted FIPs significantly improves the detection of long-cycle, hard-to-detect faults. Our approach offers a scalable solution for test strategy optimization in large-scale SoC designs and demonstrates potential for integration into downstream Electronic Design Automation workflows.

This work was supported by the Grants-in-Aid for Scientific Research 25K15042, 23K11033, 21H03411, the bilateral joint research project JSBP120237413 between JSPS (Japan Society for the Promotion of Science) and NSFC, and NSFC Grant 62311540021 and by JST SPRING, Japan Grant Number JPMJSP2162.

# 1 INTRODUCTION

With the rapid development of cloud computing and large-scale data centers, the requirements for reliability and data integrity in computing systems have continuously intensified. Silent Data Errors (SDE), a specific type of computational error, do not trigger hardware exceptions or error alarms, yet they can lead to erroneous results being returned by the system, severely affecting application correctness and system stability [1-11]. SDEs predominantly originate from time-zero defects introduced during chip manufacturing, as well as from aging-induced degradation during operational use [1]. These defects are frequently embedded within the critical timing paths of a circuit, rendering them particularly challenging to detect via conventional scan-based testing approaches [12, 13]. Consequently, functional testing remains essential for activating and identifying such elusive defects [14, 15, 16].

However, functional testing for large-scale circuits is inherently complex. The sheer number of potential activation paths renders exhaustive testing nearly impractical. To mitigate this problem, the concept of Functionally-Possible Faults (FPFs) has been introduced [28], which prioritizes faults likely to be activated under typical functional operations, thereby enabling a more targeted and efficient testing strategy. At present, mainstream approaches for FPF analysis predominantly rely on functional simulation or static structural timing analysis including path prioritization based on signal switching frequency, static analysis considering path length or delay, and propagation probability modeling [17-29]. While these methods have improved testing efficiency to a certain extent, they remain limited by excessive computational resource requirements and inadequate modeling of dynamic timing behaviors, thus posing significant challenges to the scalability and cost-effectiveness of functional testing in modern large-scale integrated circuits.

Given the inherently graphical structure of circuit netlists and physical layouts, Graph Convolutional Networks (GCNs) have emerged as a promising solution to a variety of Electronic Design Automation (EDA) challenges. GCNs have achieved notable success in several EDA stages, including operation delay prediction and hardware mapping optimization in logic synthesis, macro placement, congestion estimation, parameter optimization during placement and routing, as well as power estimation and test point analysis in verification and testing [30-36]. In practical applications, GCNs have substantially improved automation levels and prediction accuracy throughout the EDA workflow. Nevertheless, most existing GCN models are limited to combinational circuits or static single-cycle graph representations, which are insufficient for capturing the temporal dynamics inherent in fault propagation within sequential circuits. This limitation significantly impairs their effectiveness in functional fault prediction tasks.

To address the aforementioned challenges, this paper proposes a Spatio-Temporal Graph Convolutional Network (ST-GCN) framework that jointly models circuit topology and time-dependent fault propagation probabilities. The proposed framework enables rapid and quantitative prediction of long-cycle functional fault probabilities, thereby providing robust support for enhancing both the efficiency and reliability of chip testing. Unlike traditional methods that rely on extensive simulations, our approach effectively integrates both circuit structural information and the temporal evolution of signal propagation, allowing for the swift identification and quantitative assessment of functionally possible faults. Specifically, multi-cycle gate-level circuits are formulated as time-evolving graph structures, with nodes representing logic gates, edges denoting signal lines, and the temporal dimension encapsulating the dynamic variation of fault impact probabilities across cycles. By simultaneously considering both structural and timing characteristics, the ST-GCN framework facilitates comprehensive spatiotemporal modeling of the functional impacts of faults.

The proposed method supports multiple feature modeling strategies based on testability metrics and fault simulation, allowing flexible trade-offs between computational efficiency and prediction accuracy. Furthermore, the model is applicable to various test architectures, including functional testing, test point insertion (TPI), and multi-cycle logic BIST. The generalization capability of the model across different test optimization scenarios is validated through a final TPI case study.

The main contributions of this study are as follows:

- We propose the ST-GCN framework, which seamlessly integrates circuit structural topology and timing dynamics. In contrast to traditional GCN models focused solely on static, single-cycle analysis, our ST-GCN enables dynamic, long-cycle fault prediction.
- We introduce the concept of Fault Impact Probability to directly quantify and assess the impact of functionally possible faults. Additionally, we propose two feature modeling strategies that offer a flexible trade-off between computational efficiency and prediction accuracy.
- Experimental results confirm that the ST-GCN model offers significant computational advantages over traditional functional simulation methods, particularly in long-cycle functionally possible fault prediction tasks.
- Through a case study on test point selection, we systematically validate the practical effectiveness and scalability of the ST-GCN framework in integrated circuit test optimization.

The rest of this paper is organized as follows. Section 2 reviews the research background and related work. Section 3 introduces key definitions and problem modeling. Section 4 details our fault impact probability prediction framework based on Graph Convolutional Networks. Section 5 describes the experimental setup and evaluation results. Section 6 presents a case study on test point selection, and Section 7 concludes the paper.

## 2 BACKGROUND

### 2.1 Silent Data Errors (SDE) and Functional Testing

The rapid proliferation of large-scale data centers and cloud computing infrastructures has significantly increased the demands for computing system reliability and data integrity. Silent Data Errors (SDEs) are computational faults that arise during system operation and, while producing incorrect results, do not trigger observable hardware exceptions or error signals [1]. Major cloud service providers, including Google, Meta (Facebook), and Intel, have frequently observed SDEs across their vast server and chip populations [3, 4, 5]. Although SDEs do not induce system crashes, they can return corrupted data, posing a severe threat to overall system reliability.

The origins of SDEs fall into two primary categories: time-zero defects undetected during manufacturing test phases and aging-induced degradations that manifest during operational use [12, 13]. The manifestation of these faults depends not only on inherent defect characteristics but also on environmental factors such as temperature, voltage, and the timing sensitivity of the affected logic paths [13]. Defects residing within critical timing paths exhibit exacerbated impacts, whereas those in redundant or non-critical modules may only become active under specific functional states or extreme environmental conditions. Consequently, such defects are often elusive to traditional scan-based testing methods, necessitating the adoption of functional testing as a supplementary strategy to enhance defect coverage [13].

Functional testing executes circuits at operational speed over extended clock cycles, which is critical for detecting defects that only manifest under prolonged or functional operating conditions [14]. However, the combinatorial complexity of fault types and logical paths, coupled with the requirement to consider circuit behavior over numerous cycles, renders the generation of functional test patterns computationally prohibitive. Exhaustively testing all possible faults across every path is infeasible in practice. Consequently, prioritizing functionally significant faults is crucial for optimizing test efficiency and ensuring robust chip reliability.

## 2.2 Functionally-Possible Faults (FPFs)

Functionally-possible faults (FPFs) are selected from comprehensive fault models, including stuck-at, transition, delay, defect-aware, cell-aware, and gate-exhaustive faults. FPFs refer specifically to faults that can be activated and result in functional failures during normal circuit operation [18].

Mainstream identification approaches for FPFs primarily rely on functional simulation to evaluate fault impact. For instance, in [18, 29], diverse sets of input stimuli are applied to gate-level circuits to determine whether each fault alters the expected circuit outputs, thereby identifying FPFs. Additionally, as shown in [30], fault prioritization based on functional switching activity quantifies the likelihood of fault occurrence during the chip's aging process, enabling efficient selection from among path delay faults. While these simulation-driven methods yield deterministic probabilities for fault impacts, they incur substantial computational overhead due to repeated multi-cycle simulations for all input patterns, resulting in significant runtime.

Alternatively, static analysis techniques have also been widely adopted. Such methods identify potential delay faults through structural or timing analysis, often targeting the longest testable or critical timing paths [19–23]. Other studies incorporate process variations [24] or account for voltage and temperature effects [25, 26, 27] on path delays. Some recent works [30, 31, 32] focus on functionally feasible paths for more targeted analysis.

Despite notable improvements in test efficiency, existing approaches present several key limitations:

1. Excessive reliance on functional simulation leads to prohibitive computational costs, hindering scalability for large-scale SoC designs.
2. The performance of many algorithms is highly sensitive to hyperparameter settings (e.g., thresholds, similarity metrics, and path length constraints), reducing their adaptability and stability across varying circuits and test scenarios.
3. Static structural analysis fails to capture the temporal evolution of FPFs during multi-cycle, multi-path propagation, limiting its descriptive power.
4. These methods do not quantify the relationship between the impact probability of FPFs and their detection likelihood, making fault risk assessment less intuitive.

There is a compelling need for novel analysis techniques that more accurately and efficiently characterize FPFs. In particular, methods that jointly integrate circuit structural and timing information are required to achieve rapid, scalable, and robust fault modeling.

### 2.3 Application of Graph Convolution Networks in Hardware

As Moore's Law continues to advance, the complexity of chip design has increased dramatically, rendering Electronic Design Automation (EDA) tools essential for achieving efficiency, reliability, and scalability throughout the design process. In recent years, machine learning (ML) techniques have been extensively adopted in various stages of EDA, substantially improving design space exploration, performance prediction, and optimization. Notably, many entities within EDA, including circuit netlists, physical layouts, and connectivity structures, inherently possess graph-based characteristics. Consequently, Graph Convolution Networks (GCNs) have emerged as powerful tools for directly processing and modeling graph-structured data in EDA.

To date, numerous studies have proposed effective GCN-based approaches for various stages of the EDA design flow. For example, GCNs have been employed for operation delay prediction and hardware mapping optimization in logic synthesis [33], as well as layout and routing optimization in physical design [35,36]. They have also enabled early and accurate estimation of critical metrics such as power consumption in verification [34], and facilitated efficient test point selection and testability modeling for design-for-testability [38, 39]. Collectively, these advances have not only enhanced the intelligence and automation of EDA workflows but have also introduced new avenues for improving the efficiency and scalability of integrated circuit design.

Although existing GCN-based methods have achieved preliminary success in static structural modeling, most studies remain confined to single-cycle or static graph paradigms. As a result, they fail to capture the time-dependent characteristics of fault propagation during circuit operation. In the context of faults such as SDEs, which exhibit delayed and non-deterministic behavior, reliance on static circuit structures is insufficient to characterize the complex, dynamic evolution of such faults.

Therefore, developing a modeling approach that simultaneously captures both the topological structure of circuits and the temporal characteristics of signal propagation remains a significant challenge in the field of reliability systems.

In this work, we propose a Spatio-Temporal Graph Convolutional Network (ST-GCN) framework that combines circuit topology with short-cycle circuit information. By leveraging spatio-temporal feature encoding through both temporally-oriented and spatially-oriented encoders, our method effectively learns the temporal evolution of circuit features. Finally, a decoder is used to predict long-cycle fault impact probabilities. This approach supports multiple feature modeling strategies, including both testability-metric-based and fault-simulation-based methods, enabling a flexible trade-off between computational efficiency and prediction accuracy.

## 3 PROBLEM MODELING AND DEFINITIONS

This section formalizes the problem setting and introduces key definitions for the proposed ST-GCN-based rapid and efficient quantification of functionally possible fault impact probabilities in circuits.

### 3.1 Definition of Fault Impact Probability (FIP)

The evaluation of fault impact varies across different testing scenarios. In functional testing, fault impact is primarily assessed at the primary outputs (POs). For multi-cycle Logic Built-In Self-Test (LBIST), evaluation occurs at pseudo-primary outputs (PPOs). In Test Point Insertion (TPI) tasks, observation points including POs, PPOs, and inserted test points are also considered. To unify these scenarios, this work defines the FIP as the likelihood that a specific fault on a circuit signal line is detected within a given clock cycle at any observation point in the comprehensive set.

The FIP not only represents the probability of fault propagation to observation points within $t$ clock cycle, but also reflects the likelihood of detection by the applied test patterns or testing mechanisms. This provides a simple and intuitive measure of how often a fault impacts the circuit output across a range of functional behaviors. Formally, for a fault $f$ on a signal line during clock cycle $t$, the fault detection probability $FIP(f, t)$ is defined as:

$$FIP(f,t) = \frac{1}{N} \sum_{p=1}^{N} I(f,t,p,O) \tag{1}$$

where $N$ is the total number of test patterns, and the indicator function $I(f, t, p, O)$ is defined as:

$$I = \begin{cases} 1, \text{if fault } \boldsymbol{f} \text{ is observed at any observation point in the set } \boldsymbol{O} \text{ during clock cycle } \boldsymbol{t} \text{ under test pattern } \boldsymbol{p} \\ 0, \text{otherwise} \end{cases}$$

For example, consider a fault $f$ on a s given signal line with a set of test patterns $\{p_1, p_2, \ldots, p_n\}$, and let the observation point set $O$ comprise only the primary outputs. For a given clock cycle $t = 2$, each pattern $p_i$ is applied, and it is recorded whether the effect of fault $f$ is observed at any primary output. If, among 10 test patterns, 7 patterns can detect the fault, in other words, fault $f$ can be activated and propagated to the primary outputs by 7 patterns, the indicator function $I(f, 2, p, PO)$ equals 1 for these 7 patterns and 0 for the remaining 3. Thus, the FIP for fault $f$ at cycle $t$ is calculated as:

$$FIP(f,2) = \frac{1}{10} \times (1+1+1+1+1+1+1+0+0+0) = 0.7$$

This result indicates a 70% probability that fault $f$ will be detected at the primary outputs during clock cycle $t$ under the applied test patterns. A higher FIP value corresponds to a higher detection rate of fault $f$, and also indicates a greater likelihood that fault $f$ will affect the POs. The FIP thus provides an intuitive quantitative measure for FPFs. It is worth noting that the impact of a fault propagating to different POs may vary depending on system-level functionality. In this paper, we focus solely on the fault effect within a single functional circuit block. Therefore, we assume that each observed fault propagation to a PO contributes equally to the FIP metric.

### 3.2 Modeling of the Spatio-Temporal Graph (ST-Graph)

Traditional static analysis techniques are insufficient for capturing the dynamic evolution of circuit signals over time. To address this limitation, we introduce a Spatio-Temporal Graph (ST-Graph) that unifies the representation of both circuit topology and the time-varying characteristics of fault propagation. Formally, an ST-Graph is defined as $ST\text{-}Graph = G(V, A, E)$ where:

- **Node set $V$**: Each node represents a logic gate or flip-flop, with $|V| = n$. Every node $v_i \in V$ is associated with a static feature vector $H_i = \mathbb{R}^d$, where the dimensionality $d$ is determined by the gate type. These vectors characterize the influence of various gate categories on fault propagation.
- **Edge set $A$:** The edge set describes interconnections between nodes, representing signal propagation paths via an adjacency matrix $A$. Every edge $e_{ij} \in E$ is assigned a dynamic feature vector $e_{ij}^{(t)} \in \mathbb{R}^p$ at each discrete time step $t$, where $p$ denotes the dimensionality of the edge features. These time-dependent features reflect the temporal dynamics of fault propagation.
- **Time-series matrix $E$:** Over the discrete interval $[T, T + m]$, the dynamic edge features form a sequence $E = \{E^{(t)} \in \mathbb{R}^{|E| \times p} \mid t = T, \ldots, T + m\}$. Where each $E^{(t)}$ captures the comprehensive edge state at time $t$, resulting in a temporal sequence of length $m$.

This modeling approach seamlessly integrates structural and temporal information, providing a comprehensive foundation for subsequent spatio-temporal fault impact analysis.

## 3.3 Formal Problem Definition

Traditional approaches, such as static timing analysis and single-cycle GCN models, are insufficient for capturing the long-range propagation characteristics of defects with adequate accuracy and efficiency. To address these limitations, we propose a ST-GCN framework designed to facilitate efficient and accurate prediction of long-cycle FIPs.:

- $V$ and $E$ denote the set of logic-gate nodes and signal-line edges, respectively;
- $H = \{H_i\}_{i=1}^{n}$ represent the static feature vectors associated with the $n$ gates; and
- $\{E^{(t)}\}_{t=T}^{T+m}$ denote the sequence of dynamic edge-feature matrices observed over the interval $[T, T+m]$.

The objective is to design and train an ST-GCN predictor $f_{ST-GCN}$, such that, given the ST-Graph data from the previous $m$ cycles, the FIP for the subsequent $s$ cycles can be accurately predicted, as expressed by the following mapping:

$$f_{ST-GCN}: \left(V, E, H, \{E^{(t)}\}_{t=T}^{T+m}\right) \to \{\hat{Y}^{(t)}\}_{t=T+m+1}^{T+m+s} \tag{2}$$

where $\hat{Y}^{(t)}$ denotes the vector of predicted FIP values on the output signal lines of logic gates at future cycle $t$. In compact form,

$$\{\hat{Y}^{(t)}\}_{t=T+m+1}^{T+m+s} = f_{ST-GCN}\left(V, E, H, \{E^{(t)}\}_{t=T}^{T+m}; W\right) \tag{3}$$

with $W$ representing the set of learnable parameters of the network.

In summary, the problem of FIP prediction is formulated as a supervised learning task, wherein the model is trained to learn the mapping between input spatio-temporal graph features and ground-truth FIP labels obtained from fault simulation, thus enabling accurate prediction of future FIP values.

## 4 METHOD

This section introduces a method based on the ST-GCN framework for predicting the long-cycle impact probability of functionally possible faults in circuits. An overview of the proposed method is depicted in Figure 1. The proposed approach comprises two key components: (1) an ST-Graph converter for generating graph structures compatible with graph convolutional networks; (2) an ST-GCN framework equipped with a feature encoder to extract spatial and temporal features from the circuit; and a decoder for predicting future-cycle fault impact probabilities.

### 4.1 ST-Graph Converter

To leverage GCNs for joint modeling of the structural and temporal characteristics of circuits, we systematically transform the gate-level netlist into a ST-Graph. Specifically, Figure 2 presents an example that demonstrates the transformation process from a logic circuit to its corresponding graph representation. Each node is associated with a 9-dimensional one-hot encoded feature vector to indicate the gate type (e.g., AND, OR, D flip-flop). Edges encode the interconnections and signal propagation pathways among logic gates, and the overall graph structure is captured using an adjacency matrix $A$.

To construct the temporal dynamic features $E^{(t)}$ on signal lines, and to balance computation time and accuracy, we provide two methods for modeling the temporal feature sequence on signal lines.

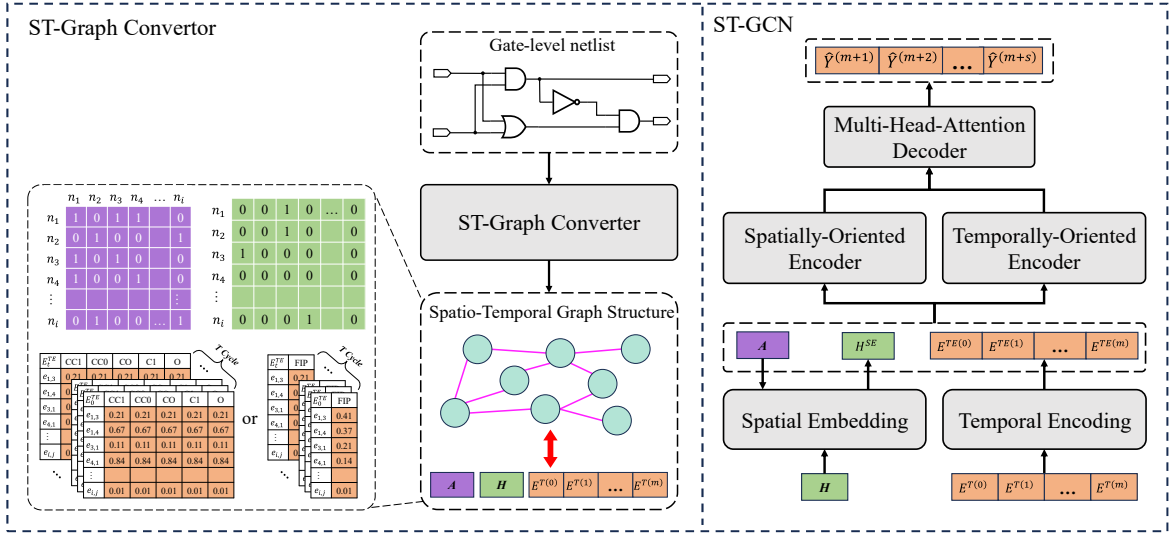

Fig. 1. Architecture of the Proposed ST-GCN-Based FIP Prediction Framework

### 4.1.1 Testability Metrics-Based Method

The temporal feature sequence for each signal line consists of controllability metrics (CC0, CC1), observability (CO), the probability of controlling a signal line to logic 1 (C1), and the probability of observing a signal line (O). These metrics are calculated using standard testability evaluation techniques, such as Sandia Controllability/Observability Analysis Program (SCOAP) and Controllability/Observability Probability (COP). The resulting temporal feature matrix is depicted in Fig. 3.1, which demonstrates the distribution of testability metrics across signal lines over time. It is worth noting that since the values of CC0, CC1, and CO range from 0 to positive infinity, min–max normalization is applied to them in advance to accelerate model convergence.

This method is based solely on the circuit's structural information, thus enabling computationally efficient and rapid construction of temporal feature sequences for gate-level circuits without the need for extensive simulations. Nevertheless, this approach has inherent limitations, particularly in handling redundant paths where multiple logic branches converge or merge. Such overlapping paths can result in inconsistencies between the estimated and actual testability, consequently degrading the predictive accuracy of the model.

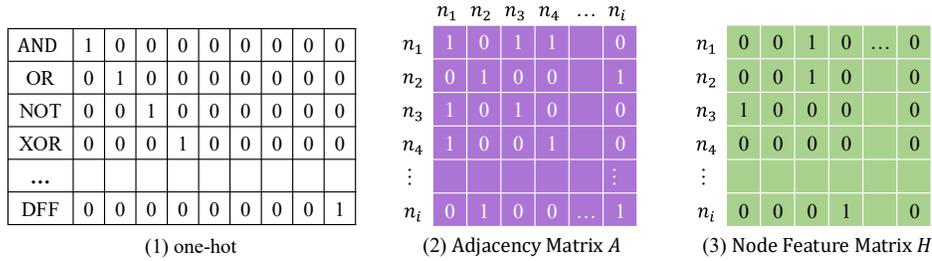

Fig. 2. Process of Transforming a Logic Circuit into a Spatial Graph Structure for ST-Graph Construction.
(1) One-hot encoding for gate types. (2) Adjacency matrix representation of circuit connectivity. (3) Node feature matrix for gates.

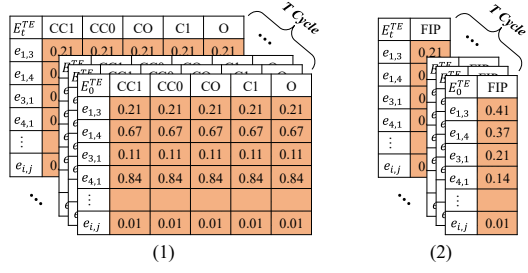

Fig. 3. Temporal Feature Sequence Construction $E^t$ for Logic Circuits:
(1) Testability Metrics-Based Methods. (2) Fault Simulation-Based Methods.

*4.1.2 Fault Simulation-Based Method:*

To address the limitations arising from path convergence, the fault simulation-based approach is adopted. Multiple sets of random or specifically designed test patterns are applied to the circuit, and multi-cycle fault simulations are conducted. For each fault on every signal line, the FIP is computed and recorded for every clock cycle. The resulting temporal feature matrix is illustrated in Fig. 3.2, providing a detailed temporal characterization of fault impacts across signal lines. While this method offers a more accurate characterization of the actual FIP and enhances the model's predictive accuracy, it incurs substantial computational overhead.

**4.2 ST-GCN**

To address the complexity of signal propagation in circuits across both spatial and temporal dimensions, we propose an ST-GCN framework that jointly models these aspects. The framework consists of four core modules: (1) spatial feature embedding and temporal feature encoding layer, (2) spatially-oriented feature encoder, (3) temporally-oriented feature encoder, and (4) spatio-temporal joint feature decoder.

*4.2.1 Spatial Feature Embedding and Temporal Feature Encoding Layer*

Initially, this layer simultaneously embeds node and edge features of the ST-Graph, thereby facilitating efficient learning of both spatial and temporal representations. Specifically, the node features $H$, the edge feature sequences $E^{(t)}$, and the adjacency matrix $A$ are provided as model inputs. The node features are projected into a unified embedding space via a linear transformation. To explicitly model the temporal dependencies inherent in the edge features, a time encoding mechanism is introduced. This mechanism generates time-specific vectors at each discrete time step, which are concatenated with the original edge features to comprehensively capture the dynamic evolution of edge attributes:

$$H_i^{SE} = W_1^{emb} H_i + W_2^{emb} \sum_{j \in \mathcal{N}(i)} H_j \tag{4}$$

$$E_{i,j}^{TE(t)} = \left[ E_{i,j}^{T(t)} || Embedding\left(E_{i,j}^{T(t)}\right) \right] \tag{5}$$

where $H_i^{SE}$ denotes the spatially encoded feature of node $i$, $E_{i,j}^{TE(t)}$ denotes the temporally encoded feature of edge $e_{i,j}$ at time step $t$, and $||$ denotes vector concatenation. Here, $\mathcal{N}(i)$ represents the set of neighboring nodes of node $i$.

*4.2.2 Spatially-Oriented Feature Encoder*

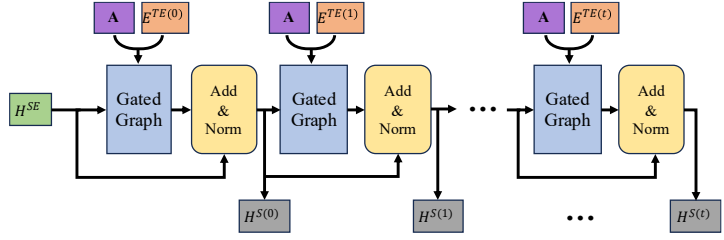

Fig. 4. Spatially-Oriented Feature Encoder Architecture

The spatially-oriented encoder models the dynamic interactions among node features within the adjacency structure using multiple layers of gated graph convolutional networks, specifically residual gated graph convolutional networks [40]. As illustrated in Figure 4, at each time step, the spatially encoded node features $H_i^{SE}$ and the temporally encoded edge features $E_{i,j}^{TE(t)}$ are jointly input into the corresponding gated graph convolutional layers. By leveraging the gating mechanism, the model effectively learns the dynamic relationships between each node and its neighbors, thereby enhancing its capability to capture complex spatial topological dependencies. Each gated graph convolutional layer incorporates residual connections and layer normalization to improve training stability and mitigate the vanishing gradient problem. The node feature update rule is defined as follows:

$$\eta_{ij}^t = Sigmoid\left(W_2^{S(t)}\left[H_i^{SE} \parallel E_{ij}^{TE(t)}\right] + W_3^{S(t)}\left[H_j^{SE} \parallel E_{ij}^{TE(t)}\right]\right) \tag{6}$$

$$H_i^{S(t)} = W_1^{S(t)} H_i^{SE} + \sum_{j \in \mathcal{N}(i)} \eta_{ij}^t \odot W_4^{S(t)}\left[H_j^{SE} \parallel E_{ij}^{TE(t)}\right] \tag{7}$$

where $\eta_{ij}^t$ denotes the gating coefficient for the edge connecting node $i$ and node $j$ at time step $t$, and $\odot$ indicates element-wise multiplication. The resulting node feature $H_i^{S(t)}$ represents the updated embedding of node $i$, which incorporates spatial interactions with its neighbors at time $t$.

Within the spatially-oriented feature encoder, the gating coefficient $\eta_{ij}^t$ adaptively modulates the strength of information propagation along each signal line (edge) from a source gate (node $j$) to a target gate (node $i$) at each time step. The computation of the gating coefficient integrates the spatial features of both the source and target logic gates, as well as the current temporal attributes of their connecting signal line. This gating mechanism allows the model to emphasize the influence of dominant signal paths in the circuit while suppressing the effects of redundant or noisy paths. By leveraging the gating coefficients, the spatial encoder can more accurately capture the local topological structure and dynamic signal behaviors in the circuit, enabling a precise characterization of how functional faults propagate along different signal lines. This process enables the model to effectively capture spatial dependencies among nodes in the graph, thereby enhancing its capacity to model complex circuit behaviors within dynamic temporal contexts.

*4.2.3 Temporally-Oriented Feature Encoder*

The temporally-oriented encoder employs a multi-head attention mechanism within a graph Transformer module [41] to achieve fine-grained representation learning of graph data at each discrete time step, as illustrated in Figure 5. At each time step $t$, the spatially encoded node features $H_i^{SE}$, temporal edge features $E_{i,j}^{TE(t)}$, and adjacency matrix $A$ are fed into the graph Transformer layer. Leveraging the global edge attention mechanism, the module effectively captures dependencies between nodes and edges across the entire graph, thereby significantly enhancing node feature representations. Each graph Transformer layer also incorporates residual connections and layer normalization to promote stable training and mitigate the vanishing gradient problem. The node feature update is as follows:

$$s_{ij}^t = \frac{(W_3^{T(t)} H_i^{SE})^\top (W_4^{T(t)} H_j^{SE} + W_5^{T(t)} E_{ij}^{TE(t)})}{\sqrt{d}} \tag{8}$$

$$\alpha_{ij}^t = \frac{\exp(s_{ij}^t)}{\sum_{k \in N(i)} \exp(s_{ik}^t)} \tag{9}$$

$$H_i^{T(t)} = W_1^{T(t)} H_i^{SE} + \sum_{j \in \mathcal{N}(i)} \alpha_{ij}^t \left( W_2^{T(t)} H_j^{SE} + W_5^{T(t)} E_{ij}^{TE(t)} \right) \tag{10}$$

where $\alpha_{ij}^t$ denotes the attention weight between node $i$ and node $j$ at time step $t$, $d$ is the dimension of the node feature embeddings, and $H_i^{T(t)}$ represents the temporally encoded feature vector of node $i$ at time step $t$.

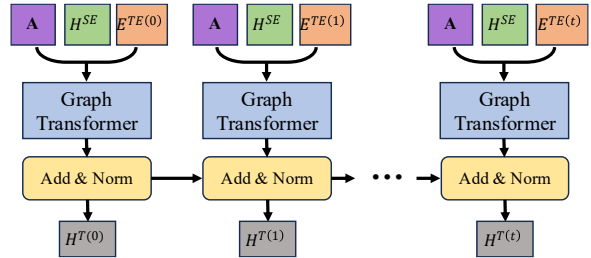

Fig. 5. Temporally-Oriented Feature Encoder Architecture

The attention coefficient $\alpha$ reflects the relative temporal importance of information from different neighboring nodes at the current time step. It enables the model to effectively distinguish which neighboring nodes' historical and current behaviors are most relevant for predicting the fault impact probability in future cycles. By assigning higher attention coefficients to nodes and edges that exert greater influence in temporal propagation, the encoder ensures that the model captures key long-range temporal dependencies and propagation patterns, thereby improving the prediction accuracy. In addition, the softmax normalization of $\alpha_{ij}^t$ across all neighboring nodes guarantees that the model maintains sensitivity to local signal dynamics while also considering the global temporal context of the entire circuit. This process enables the model to effectively capture temporal dependencies among nodes and edges, thereby providing enriched temporal feature representations that are critical for accurately modeling dynamic fault propagation over time.

Although both the gating coefficient $\eta$ and the attention weight $\alpha$ are used to aggregate information from neighboring logic gates and signal lines, they differ significantly in their computational mechanisms and modeling focus. The gating coefficient, implemented via a sigmoid activation, independently modulates the information flow on each signal line, emphasizing the presence and strength of connections between gates. This enables the selective enhancement or suppression of information along specific paths. In contrast, the attention weight is computed using a softmax normalization, distributing the focus among all neighboring logic gates and highlighting the most relevant nodes, which is particularly effective for capturing global dependencies. The joint use of these two mechanisms allows the model to suppress irrelevant local paths while simultaneously modeling global temporal dependencies, thereby enhancing the overall representational power and predictive accuracy of the network, as shown in Fig. 6.

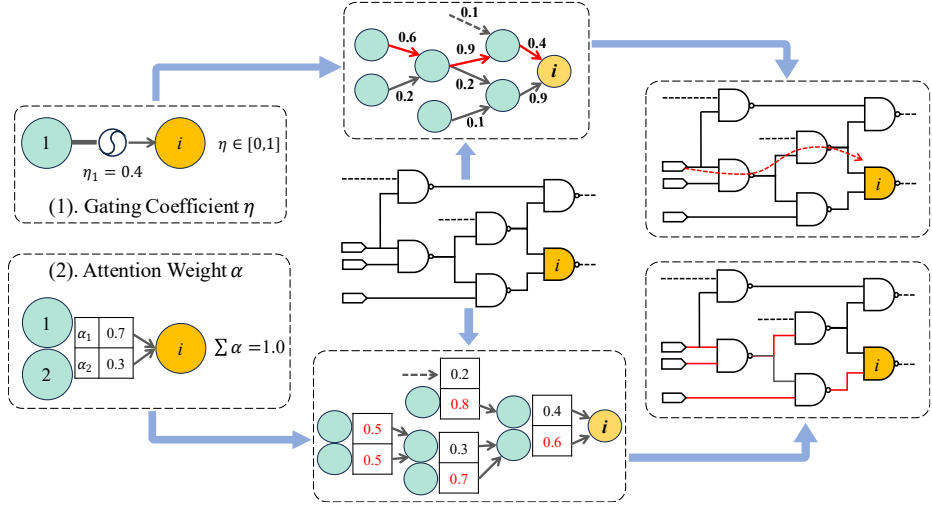

Fig. 6. Comparison of Gating and Attention Mechanisms for Information Aggregation in ST-Graph

*4.2.4 Spatio-Temporal Joint Feature Decoder*

Finally, the spatio-temporal joint feature decoder aggregates the node embeddings generated by both the spatially oriented and temporally oriented encoders. At each time step, the embeddings from the spatial encoder $H_i^{S(t)}$ and the temporal encoder $H_i^{T(t)}$, are combined to produce a joint spatio-temporal feature sequence for each node:

$$H^{ST} = \{H^{S(0)} + H^{T(0)}, H^{S(1)} + H^{T(1)}, \ldots, H^{S(t)} + H^{T(t)}\} \tag{11}$$

The decoder processes these joint features to predict the FIP for the circuit. At each time step, the joint feature vector $H^{ST}$ first goes through an attention-based aggregation. This step extracts the most important information from global spatio-temporal dependencies. The aggregated vector is then projected into the output space through a linear transformation. Finally, a sigmoid activation function maps the result to the range [0,1], so it can be interpreted directly as a probability:

$$H_i^{out} = Sigmoid\left(\left(softmax\left(\frac{W_2^{out} H_i^{ST}(W_3^{out} H_i^{ST})^\top}{\sqrt{d}}\right) W_1^{out} H_i^{ST}\right) W_4^{out}\right) \tag{12}$$

where $H_i^{ST}$ denotes the aggregated spatio-temporal feature vector for node $i$, and $H_i^{out}$ $i$ predicted FIP vector for the output signal line associated with node $i$. The matrices $W^{out}$ represent the learnable parameters of the decoder.

The attention-based decoder keeps node-to-node interactions in the prediction stage. This design has two benefits: (1) the attention mechanism enhances the model's sensitivity to long-range dependencies, mitigating the loss of cross-node and cross-time information that can occur in direct regression; and (2) maintaining architectural consistency with the gating and attention mechanisms used in the encoder promotes coherent information flow and improves the efficiency of feature utilization. The ablation study in Section 5.4 further validates this design choice.

## 5 EXPERIMENTAL RESULTS

We conducted experiments on ISCAS'89 benchmark circuits to substantiate the effectiveness and efficiency of the proposed ST-GCN framework for FIP prediction. Specifically, we assessed the performance of each corresponding model in terms of both prediction accuracy quantified as the discrepancy between predicted and simulated FIP values and computational efficiency, measured by the time required for ST-Graph data conversion and model inference.

In this section, we explain the construction of the dataset, the experimental setup, and the analysis of the results.

### 5.1 Dataset and Model

To comprehensively evaluate the FIP prediction performance of the proposed ST-GCN framework under different input features and dataset coverage, we designed and constructed multiple datasets based on ISCAS'89 benchmark circuits as follows:

First, for each gate-level netlist, two types of ST-Graphs were generated using the ST-Graph Converter: one based on structural testability metrics, and the other based on ground-truth fault simulation results, covering stuck-at, transition, and delay faults. Each type of ST-Graph spans 20 clock cycles. For simulation-based ST-Graphs, 10,000 random test patterns were applied to each fault, and fault simulation was performed for 20 consecutive clock cycles per pattern, with the observation point set consisting only of POs (functional simulation). The ground-truth FIP for each signal line at every cycle was then recorded.

To augment the dataset, we adopted a sliding window approach, partitioning the 20-cycle ST-Graphs into multiple shorter ST-Graph samples of lengths 10 and 15. Specifically, the FIP or testability metrics from the most recent 5 cycles were used as input to predict the FIP for the subsequent 5 or 10 cycles, thereby generating corresponding ST-Graph samples for model training and evaluation.

For dataset partitioning, all benchmark circuits were sorted in ascending order of size, and two sampling strategies were adopted for training set construction: (1) uniform sampling, where every other circuit is selected as a training sample; and (2) sparse sampling, where every two circuits are selected as training samples.

Following this procedure, we constructed a total of eight datasets, encompassing different input feature types (testability metrics or simulation-based FIP), training set sampling strategies (uniform or sparse), and prediction horizons (5-cycle or 10-cycle prediction), thus providing comprehensive experimental coverage. For each combination of input type (simulation-based FIP or testability metrics) and prediction horizon (5 cycles or 10 cycles), an independent ST-GCN model is trained and evaluated. For simplicity, the models are referred to as FIP-5, FIP-10, TM-5, and TM-10, corresponding to simulation-based FIP or testability-metric-based inputs with 5- or 10-cycle prediction horizons, respectively. For different training set sampling strategies, the suffixes 'U' (Uniform) and 'S' (Sparse) are appended as needed.

### 5.2 Experimental Setup

For all models, mean squared error (MSE) was uniformly adopted as the loss function to optimize the network parameters, and training was performed for 200 epochs using the Adam optimizer with an initial learning rate of 0.05. Upon completion of training, only root mean squared error (RMSE) and mean absolute error (MAE) were employed to evaluate the predictive performance of the models. RMSE restores the error to the same unit as the original data and is more sensitive to larger deviations, while MAE, owing to its linear penalty, offers greater robustness to outliers. Together, these two metrics comprehensively reflect the accuracy and stability of the models in long-cycle, multi-step prediction tasks. The definitions of these metrics are as follows:

$$MSE = \frac{1}{n \times s} \sum_{i=1}^{n} \sum_{t=1}^{s} (y_{i,t} - \hat{y}_{i,t})^2, \quad RMSE = \sqrt{MSE}, \quad MAE = \frac{1}{n \times s} \sum_{i=1}^{n} \sum_{t=1}^{s} |y_{i,t} - \hat{y}_{i,t}| \tag{13}$$

where $y_{i,j}$ and $\hat{y}_{i,j}$ denote the true and predicted values for the $i$-th sample at the $j$-th future time step, respectively.

The experimental hardware configuration consisted of an Intel i9-14900 KF processor and an NVIDIA GeForce RTX4090 GPU with 24GB of memory. Both the model's training, inference, and data processing were carried out in this environment to ensure the stability and reproducibility of the results.

### 5.3 Evaluation Results

*5.3.1 Computational Efficiency Analysis*

The computational efficiency of the ST-GCN framework was systematically evaluated by measuring the time required for ST-Graph data conversion, as well as the time for model training and inference under different task models.

- Figure 7.1 compares the computational overhead of simulation-driven and testability-metric-based methods during ST-Graph data conversion, with the conversion window fixed at 5 clock cycles. The experimental results show that the computational time required by traditional fault simulation methods increases exponentially with circuit size, whereas the testability-metric-based approach exhibits only linear growth and remains significantly lower than that of fault simulation throughout.
- Figure 7.2 presents the maximum training time required for a single epoch across all models. It can be observed that as the circuit size increases, the maximum training time for the models also grows accordingly. However, it is important to note that the training process only needs to be performed once, and the trained model can be efficiently utilized for subsequent prediction tasks.
- Figure 7.3 further compares the inference times of the FIP-5 and FIP-10 models on both CPU and GPU platforms (excluding data conversion time). The results show that, on the CPU platform, the average inference time per 10,000 logic gates is approximately 52 seconds, while on the GPU platform, this time is reduced to about 4 seconds. Moreover, when extending the prediction horizon from 5 to 10 clock cycles, the increase in inference time is not significant, indicating that the proposed method maintains excellent scalability and high efficiency even for longer prediction windows.

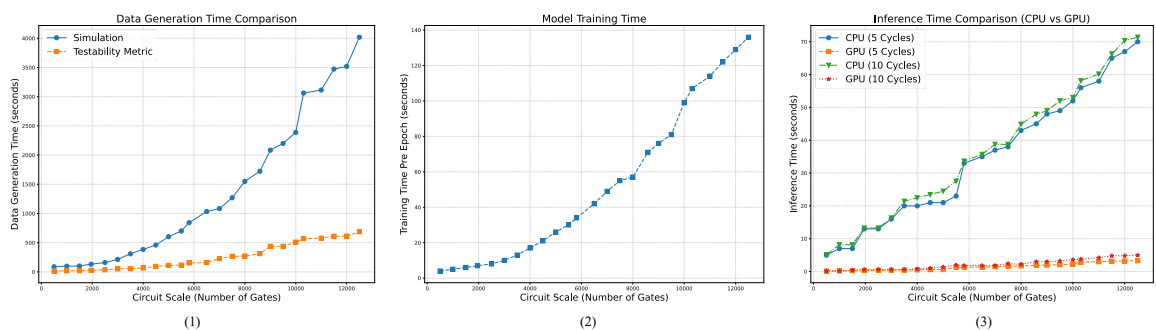

Fig. 7. Computational Efficiency of ST-GCN: (1) ST-Graph Data Conversion, (2) Training, and (3) Inference Time versus Circuit Scale.

Compared to traditional simulation-based methods, the proposed approach demonstrates significant advantages in computational efficiency, substantially reducing the time required for conventional simulations. This makes it particularly suitable for accelerating the evaluation of functionally possible faults in large-scale circuits.

Table 1: ST-GCN Prediction Accuracy Using Simulation-Based FIP

| Model | | FT-5-U | | FT-10-U | | FT-5-S | | FT-10-S | |
|---|---|---|---|---|---|---|---|---|---|
| Evaluation Function | | RMSE | MAE | RMSE | MAE | RMSE | MAE | RMSE | MAE |
| Circuit | s298 | 0.0045 | 0.0032 | 0.0057 | 0.0044 | 0.0037 | 0.0029 | 0.0149 | 0.0080 |
| | s344 | 0.0505 | 0.0164 | 0.0520 | 0.0176 | 0.0668 | 0.0197 | 0.0564 | 0.0196 |
| | s349 | 0.0503 | 0.0146 | 0.0552 | 0.0165 | 0.0559 | 0.0146 | 0.0543 | 0.0178 |
| | s382 | 0.0672 | 0.0228 | 0.0656 | 0.0221 | 0.1345 | 0.0563 | 0.1566 | 0.0657 |
| | s386 | 0.0864 | 0.0418 | 0.0857 | 0.0448 | 0.0697 | 0.0353 | 0.0731 | 0.0424 |
| | s420 | 0.0632 | 0.0088 | 0.0632 | 0.0094 | 0.0662 | 0.0089 | 0.0750 | 0.0128 |
| | s444 | 0.0798 | 0.0122 | 0.0937 | 0.0164 | 0.0433 | 0.0096 | 0.0495 | 0.0135 |
| | s510 | 0.0882 | 0.0174 | 0.0881 | 0.0196 | 0.1028 | 0.0253 | 0.1092 | 0.0308 |
| | s641 | 0.0730 | 0.0247 | 0.0825 | 0.0295 | 0.0845 | 0.0276 | 0.0871 | 0.0294 |
| | s713 | 0.0958 | 0.0384 | 0.1141 | 0.0469 | 0.1896 | 0.0771 | 0.1865 | 0.0792 |
| | s820 | 0.0939 | 0.0379 | 0.0941 | 0.0404 | 0.1022 | 0.0428 | 0.0937 | 0.0460 |
| | s832 | 0.0905 | 0.0363 | 0.0875 | 0.0374 | 0.0983 | 0.0412 | 0.0879 | 0.0439 |
| | s838 | 0.0446 | 0.0063 | 0.0447 | 0.0072 | 0.0467 | 0.0064 | 0.0531 | 0.0098 |
| | s953 | 0.0790 | 0.0258 | 0.0806 | 0.0269 | 0.0554 | 0.0199 | 0.0494 | 0.0195 |
| | s1238 | 0.0798 | 0.0391 | 0.0855 | 0.0394 | 0.0727 | 0.0356 | 0.0700 | 0.0354 |
| | s1488 | 0.0910 | 0.0378 | 0.0847 | 0.0401 | 0.0806 | 0.0358 | 0.0789 | 0.0407 |
| | s5378 | 0.0508 | 0.0113 | 0.0505 | 0.0119 | 0.0501 | 0.0099 | 0.0534 | 0.0126 |
| | s9234 | 0.0836 | 0.0294 | 0.0852 | 0.0322 | 0.0834 | 0.0296 | 0.0846 | 0.0310 |
| Average | | 0.0707 | 0.0236 | 0.0733 | 0.0257 | 0.0781 | 0.0277 | 0.0796 | 0.0310 |

Table 2: ST-GCN Prediction Accuracy Using Testability Metrics

| Model | | TM-5-U | | TM-10-U | | TM-5-S | | TM-10-S | |
|---|---|---|---|---|---|---|---|---|---|
| Evaluation Function | | RMSE | MAE | RMSE | MAE | RMSE | MAE | RMSE | MAE |
| Circuit | s298 | 0.0687 | 0.0229 | 0.0813 | 0.0230 | 0.0825 | 0.0204 | 0.0858 | 0.0204 |
| | s344 | 0.0760 | 0.0289 | 0.1007 | 0.0311 | 0.1188 | 0.0392 | 0.1236 | 0.0347 |
| | s349 | 0.0666 | 0.0202 | 0.0918 | 0.0243 | 0.1009 | 0.0279 | 0.0951 | 0.0247 |
| | s382 | 0.0947 | 0.0309 | 0.1044 | 0.0345 | 0.1638 | 0.0609 | 0.2179 | 0.0864 |
| | s386 | 0.0938 | 0.0490 | 0.1028 | 0.0492 | 0.1187 | 0.0505 | 0.1144 | 0.0494 |
| | s420 | 0.1072 | 0.0386 | 0.1153 | 0.0390 | 0.1383 | 0.0464 | 0.1393 | 0.0438 |
| | s444 | 0.0803 | 0.0131 | 0.1077 | 0.0194 | 0.1209 | 0.0228 | 0.0957 | 0.0159 |
| | s510 | 0.0954 | 0.0213 | 0.1038 | 0.0193 | 0.1326 | 0.0330 | 0.1435 | 0.0292 |
| | s641 | 0.1065 | 0.0451 | 0.1105 | 0.0438 | 0.1372 | 0.0549 | 0.1368 | 0.0503 |
| | s713 | 0.1084 | 0.0493 | 0.1249 | 0.0583 | 0.2107 | 0.0898 | 0.2072 | 0.0832 |
| | s820 | 0.1019 | 0.0391 | 0.1101 | 0.0428 | 0.1174 | 0.0369 | 0.1161 | 0.0403 |
| | s832 | 0.0969 | 0.0375 | 0.1035 | 0.0407 | 0.1094 | 0.0339 | 0.1089 | 0.0382 |
| | s838 | 0.1003 | 0.0340 | 0.1064 | 0.0338 | 0.1286 | 0.0407 | 0.1274 | 0.0389 |
| | s953 | 0.0664 | 0.0210 | 0.0936 | 0.0269 | 0.0836 | 0.0265 | 0.0886 | 0.0233 |
| | s1238 | 0.1157 | 0.0689 | 0.1262 | 0.0668 | 0.1277 | 0.0645 | 0.1360 | 0.0677 |
| | s1488 | 0.0905 | 0.0377 | 0.0987 | 0.0416 | 0.1204 | 0.0417 | 0.1173 | 0.0410 |
| | s5378 | 0.0724 | 0.0152 | 0.0802 | 0.0188 | 0.0854 | 0.0206 | 0.0888 | 0.0152 |
| | s9234 | 0.0867 | 0.0271 | 0.0970 | 0.0317 | 0.1038 | 0.0293 | 0.1172 | 0.0308 |
| Average | | 0.0905 | 0.0333 | 0.1033 | 0.0358 | 0.1223 | 0.0411 | 0.1255 | 0.0407 |

*5.3.2 Prediction Accuracy Analysis*

This section systematically evaluates the prediction accuracy of the proposed model under varying training set configurations, input feature types, and prediction horizons. Tables 1 and 2 present the performance of the ST-GCN models trained with different input features simulation-based FIP and testability metrics, respectively under both uniform and sparse training set sampling strategies.

Table 1 demonstrates that the model consistently achieves low RMSE and MAE across both uniform and sparse training sets for 5-cycle and 10-cycle prediction tasks, indicating high predictive accuracy and stability. While the error metrics exhibit a moderate increase as the prediction horizon extends, the overall variation remains limited, underscoring the model's robustness in long-cycle FIP prediction.

In contrast, Table 2 reports generally higher RMSE and MAE values for models trained with testability metrics-based features compared to those utilizing simulation-based FIP, reflecting a trade-off between computational efficiency and prediction accuracy. Notably, the testability metrics-based model exhibits improved performance under uniform training set conditions, indicating that sufficient and balanced data contribute to enhanced generalization. Moreover, while the choice of training set configuration exerts negligible influence on the predictive accuracy of the simulation-based model, the testability metrics-based model benefits significantly from uniform sampling in terms of stability and precision. We observe from the results of the MT-5-S and MT-10-S models that the RMSE increases while the MAE decreases. This is attributed to an increase in the number of outliers in certain circuits, accompanied by an overall reduction in the average error level. Specifically, testability metrics are more susceptible to the influence of a small subset of circuit structures, which leads to a rise in the number of outliers over longer time horizons; however, the overall fluctuation in error remains limited.

In summary, the experimental results substantiate the efficiency and accuracy of the proposed ST-GCN framework for FIP prediction. Specifically, in large-scale circuit scenarios, the method maintains high predictive accuracy while exhibiting robust long-cycle stability and strong generalization capability.

## 5.4 Ablation Study

To validate the effectiveness and necessity of the key modules in the proposed ST-GCN framework, we conduct an ablation study by selectively removing or replacing core components and evaluating the resulting performance changes under the same dataset and training configuration. The experimental design is as follows:

- Time Encoding: Removing the time encoding mechanism while retaining the original edge feature sequences as inputs, to assess the contribution of explicitly modeling temporal information to capturing multi-cycle dependencies and improving prediction accuracy;
- Only-Spatial Encoder: Retaining only the spatial encoder branch (gated GCN-based topological feature extraction) and removing the temporal encoder branch;
- Only-Temporal Encoder: Retaining only the temporal encoder branch (multi-head attention-based temporal dependency modeling) and removing the spatial encoder branch.
- Decoder Variants: Replacing the original "spatial–temporal feature concatenation/summation + projection" decoder with a three-layer MLP regression head, to investigate the effect of different feature fusion and mapping strategies on prediction stability and long-term trend modeling.

Table 3 reports the ablation study results on the ISCAS'89 benchmark circuits, using the FIP-10-U model as the baseline.

| Model Variant | RMSE | MAE | ΔRMSE vs. Full | ΔMAE vs. Full |
|---|---|---|---|---|
| Full FIP-10-U Model | 0.0733 | 0.0257 | – | – |
| w/o Time Encoding | 0.0816 | 0.0298 | +11.3% | +15.9% |
| Only-Spatial | 0.0895 | 0.0332 | +22.1% | +29.2% |
| Only-Temporal | 0.0785 | 0.0264 | +7.1% | +2.6% |
| MLP Decoder | 0.0752 | 0.0274 | +2.6% | +6.6% |

Several key observations can be drawn from Table 3:

- Removing the time encoding mechanism (w/o Time Encoding) leads to a notable increase in both RMSE and MAE, indicating that explicitly incorporating temporal position vectors is beneficial for capturing long-term dependencies and enhancing prediction accuracy.
- Retaining only a single branch (Only-Spatial or Only-Temporal) results in a substantial drop in performance. The Only-Spatial variant exhibits the largest RMSE increase (+22.1%), highlighting the critical role of temporal modeling in capturing multi-cycle signal propagation patterns; the Only-Temporal variant also suffers accuracy degradation, suggesting that spatial topological information is equally indispensable.
- Replacing the original decoder with an MLP causes a slight performance drop, but the degradation is relatively minor. This implies that while the decoder design influences the final performance, its impact is less pronounced compared to the spatial–temporal joint modeling in the encoder stage.

Overall, the complete ST-GCN framework consistently delivers the best performance across all ablation settings, confirming the complementary and indispensable roles of its constituent modules in long-cycle FIP prediction.

## 6 CASE STUDY ON TEST POINT SELECTION

This section aims to systematically demonstrate the practical effectiveness and value of our proposed ST-GCN prediction method within the test point insertion (TPI) process. Considering that test point selection strategies are not the primary focus of this work, a simple greedy algorithm was employed to select observation points in the circuit, with D flip-flops (DFFs) serving as candidate observation points, as shown in Fig. 8.1.

During the observation point selection process, we utilize the trained TM-10-U model to predict the FIP of the circuit before and after observation point insertion. First, the target circuit is converted into a testability-metric-based ST-Graph using the ST-Graph Converter. Then, the TM-10-U model predicts the FIP for the subsequent 10 clock cycles, capturing the temporal evolution of FIP across different signal lines. By analyzing these trends, a class of "cycle-sensitive faults" can be identified these faults exhibit relatively low FIP values during the initial cycles, but show a significant increase as the cycles progress. This phenomenon indicates that such faults are difficult to detect effectively in the early stages of functional testing and require the application of extended temporal test vectors to enhance their detection probability.

As shown in Fig. 8.2, In the selection procedure, the algorithm iteratively evaluates each unassigned DFF by temporarily designating it as an observation point, reapplying the TM-10-U model to predict the FIP, and calculating the reduction in the number of cycle-sensitive faults. At each iteration, the DFF that results in the greatest reduction is selected as the next observation point. This process is repeated until either the predefined observation point limit is reached or no further reduction in cycle-sensitive faults can be achieved, thus determining the final set of observation points.

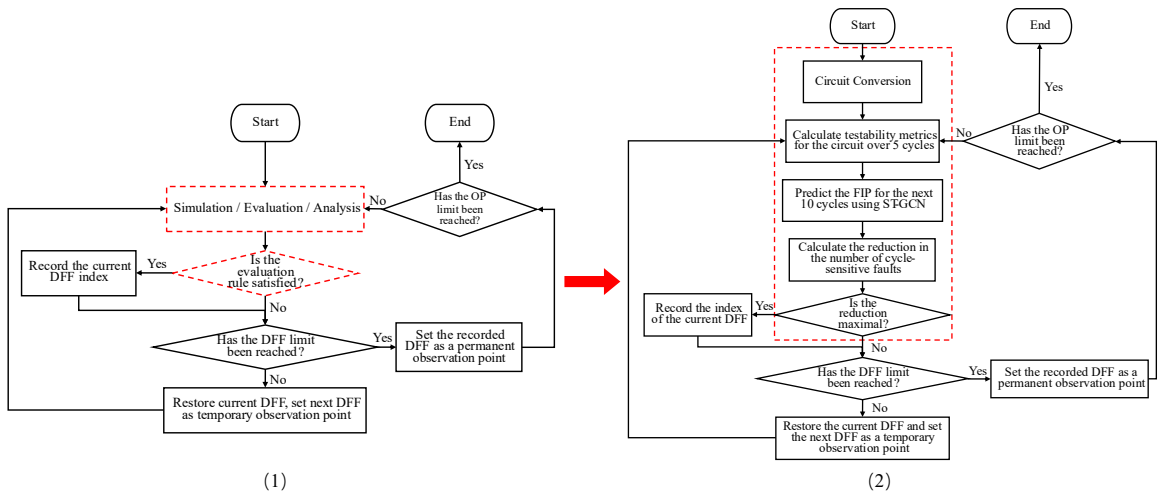

Fig. 8. Model-Assisted Greedy Observation Point Selection Flow
(1) Greedy Observation Point Selection Flow (2) ST-GCN-Based Test Point Insertion Flow.

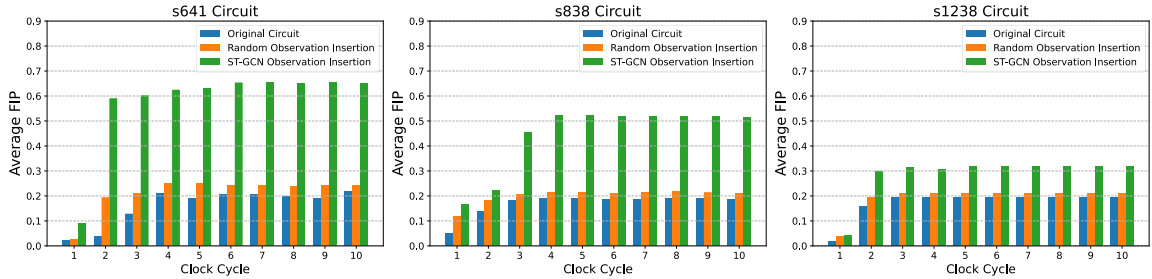

Fig. 9. Comparison of Average FIP Across Clock Cycles for s641, s838, and s1238 Benchmark Circuits

For experimental validation, the maximum number of observation points is set to 2% of the total DFFs in the circuit, and results are compared with a baseline approach in which observation points are randomly assigned. Figure 9 presents the average FIP across 10 clock cycles for the selected benchmark circuits. The findings indicate that the ST-GCN-based insertion method rapidly increases the average FIP within the first 4 clock cycles, thereby demonstrating the effectiveness and practicality of the proposed strategy in enhancing early fault detection.

This case study validates the utility of the ST-GCN model for precise test point deployment and further demonstrates the feasibility of integrating the method into existing EDA tools without necessitating complex modifications.

## 7 CONCLUSIONS

In this paper, we have proposed a ST-GCN framework to efficiently and accurately predict the functionally possible fault impact probability (FIP) in large-scale sequential circuits. Our proposed method achieves, for the first time, unified modeling of both circuit topology and the temporal dynamics of fault propagation, enabling rapid evaluation of long-cycle FIP without the need for extensive simulation. To meet diverse application requirements, two feature modeling strategies based on structural testability metrics and fault simulation are introduced, which enable a flexible trade-off between computational efficiency and prediction accuracy, demonstrating strong engineering applicability.

Experimental results on the ISCAS'89 benchmark circuits demonstrate that our proposed ST-GCN framework can efficiently predict long-cycle fault impact probabilities with minimal loss of accuracy. In particular, inference speed on the GPU platform is significantly improved, fully meeting the practical demands of industrial-scale SoC testing. The framework is also readily applicable to downstream EDA tasks, such as test compression, critical path analysis, and testability evaluation, highlighting its flexibility and scalability.

In future work, we will focus on the joint evaluation of faults, primary outputs (POs), and functional behaviors to quantitatively assess the impact of faults on both circuit functionality and system-level operation. By establishing a comprehensive bottom-up analysis framework that traces fault effects from their origin to system functions, we aim to enable a holistic assessment of Silent Data Errors (SDEs) at the system level, thereby providing data-driven insights and theoretical foundations for system-level reliability optimization.